\documentclass{article}

\usepackage{PRIMEarxiv}

\usepackage[utf8]{inputenc} % allow utf-8 input
\usepackage[T1]{fontenc}    % use 8-bit T1 fonts
\usepackage{hyperref}       % hyperlinks
\usepackage{url}            % simple URL typesetting
\usepackage{booktabs}       % professional-quality tables
\usepackage{amsfonts}       % blackboard math symbols
\usepackage{nicefrac}       % compact symbols for 1/2, etc.
\usepackage{microtype}      % microtypography
\usepackage{lipsum}
\usepackage{fancyhdr}       % header
\usepackage{graphicx}       % graphics
\usepackage{amsmath}
\usepackage{amssymb}
\usepackage{bbm}
\usepackage{multirow}
\usepackage{colortbl}
\usepackage{gensymb}
\graphicspath{{media/}}     % organize your images and other figures under media/ folder

\title{Toward Fair Facial Expression Recognition with Improved Distribution Alignment
}

\author{Mojtaba Kolahdouzi, Ali Etemad \\
Dept. ECE and Ingenuity Labs Research Institute, Queen's University, Kingston, Canada\\
}

\begin{document}
\newcommand{\E}[1]{\underset{#1}{\textbf{E}}}
\maketitle

\begin{abstract}
We present a novel approach to mitigate bias in facial expression recognition (FER) models. Our method aims to reduce sensitive attribute information such as gender, age, or race, in the embeddings produced by FER models. We employ a kernel mean shrinkage estimator to estimate the kernel mean of the distributions of the embeddings associated with different sensitive attribute groups, such as young and old, in the Hilbert space. Using this estimation, we calculate the maximum mean discrepancy (MMD) distance between the distributions and incorporate it in the classifier loss along with an adversarial loss, which is then minimized through the learning process to improve the distribution alignment. Our method makes sensitive attributes less recognizable for the model, which in turn promotes fairness. Additionally, for the first time, we analyze the notion of \textit{attractiveness} as an important sensitive attribute in FER models and demonstrate that FER models can indeed exhibit biases towards more attractive faces. To prove the efficacy of our model in reducing bias regarding different sensitive attributes (including the newly proposed attractiveness attribute), we perform several experiments on two widely used datasets, CelebA and RAF-DB. The results in terms of both accuracy and fairness measures outperform the state-of-the-art in most cases, demonstrating the effectiveness of the proposed method.
\end{abstract}

% keywords can be removed

\section{Introduction}

Facial expressions are widely recognized as one of the primary means of conveying human emotions \cite{9953876, jaques2016understanding}. Consequently, the problem of automatic facial expression recognition (FER) has been extensively investigated in the literature \cite{9894072}. Given the advancements in machine learning, especially deep learning, numerous data-driven solutions have been recently proposed to tackle this problem \cite{9666986, Najafzadeh_2023_WACV}. 
The majority of these approaches require access to large, labeled FER datasets.
One of the primary drawbacks of data-driven methods developed to address the FER problem is their vulnerability to various forms of biases in the dataset, which can lead to discrimination against certain demographic groups \cite{10.1007/978-3-030-65414-6_35}. 
% Such biases cause machine learning algorithms to make unfair decisions that could exhibit a preference for specific demographic groups. 
For instance, well-known datasets such as RAF-DB~\cite{li2017reliable}, which are frequently used for training FER models, exhibit skewed distributions with respect to sensitive attributes like age and gender~\cite{Stoychev2022TheEO}. These imbalanced distributions in the training data often result in output distributions from the machine learning algorithms that favor the majority group. 
% For instance, it has been demonstrated in \cite{?} that some FER models recognize happiness more accurately in women than in men. 
Due to the severe consequences that such biases in FER systems can cause, addressing bias in FER has recently gained increased attention in the area \cite{Zeng_2022_CVPR}. 

\begin{figure}[]
  \centering
  \includegraphics[width=0.5\linewidth]{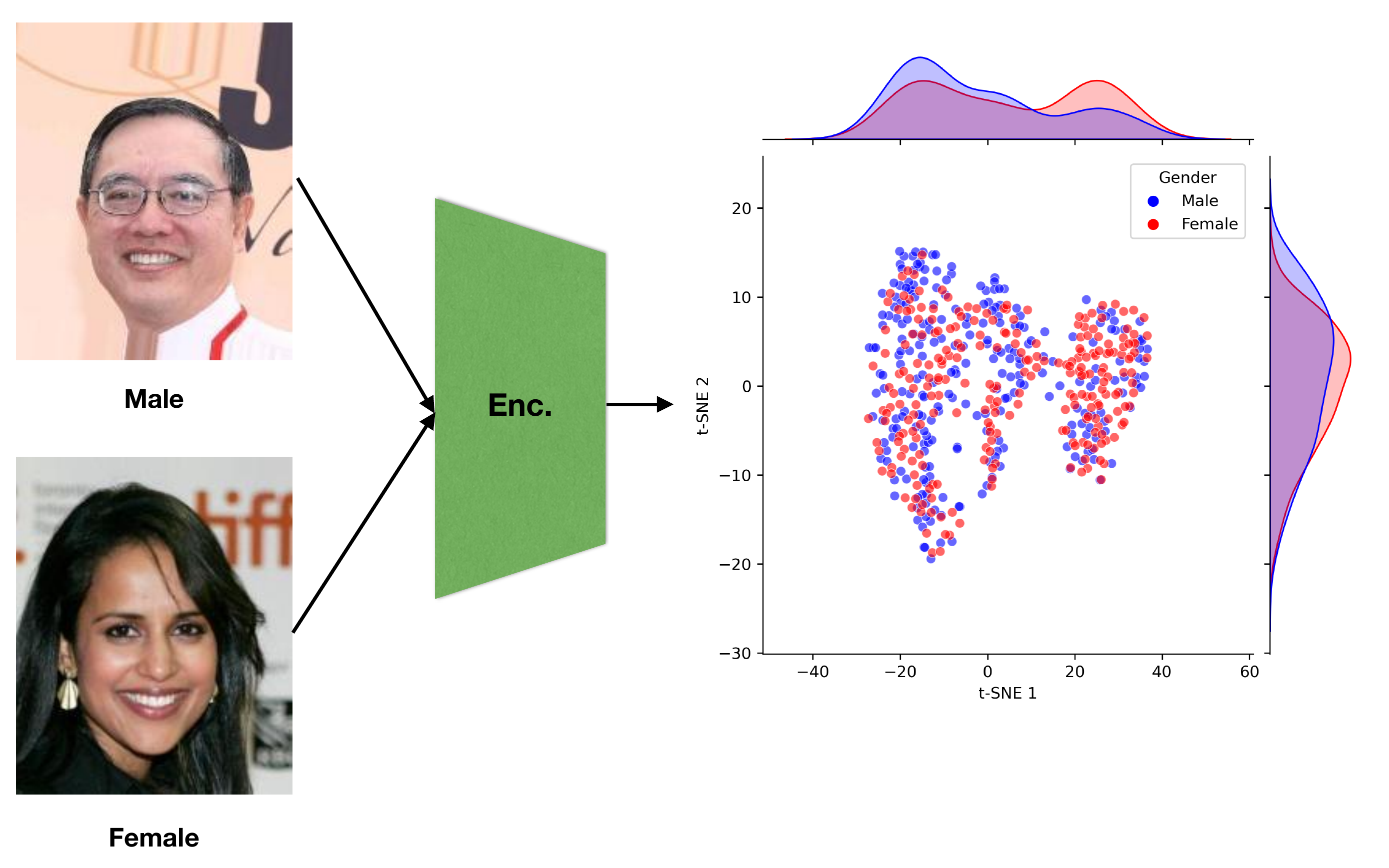}
  \caption{The difference in conditional distributions given two groups of gender as a sensitive attribute.}
  % of gender as a sensitive attribute.}
  % Visualization of the conditional distribution difference}
\label{fig:teaser}
\end{figure}

It has been previously shown that FER methods may learn representations that contain sensitive attribute information beyond the intended facial expressions alone \cite{pmlr-v115-jiang20a, Liu_2018_CVPR}. This latent information can in turn be exploited by classifiers to aid with the decision-making process, which can influence the fairness of the outcome. To further motivate this discussion, we show an example in Figure~\ref{fig:teaser} where the embeddings obtained from a ResNet18 expression classifier for male and female input samples have been processed (using t-SNE \cite{JMLR:v9:vandermaaten08a}) to generate the depicted distributions. Here, we observe a considerable shift between these distributions, which indicates that gender-related information can indeed be used to adversely impact the fairness of the classifier for the detection of particular facial expressions.
% This is in line with prior works such as \cite{??} which have demonstrated that some FER models recognize happiness more accurately in women than in men. 
Similarly, it has been demonstrated in prior works such as~\cite{dominguez2023gender} that some FER models recognize happiness more accurately in women than in men. Consequently, bias-dispelling methods have been proposed in the context of FER to enhance fairness by eliminating sensitive attribute information from the learned embeddings \cite{Alvi_2018_ECCV_Workshops, 4bf20d93df4f4b4b8496afde33993ac0}. These approaches have relied on a number of different strategies such as adversarial models \cite{9163294, 10.1145/3278721.3278779}, data-level methods \cite{NIPS2017_9a49a25d, 9042231}, and statistical techniques \cite{pmlr-v115-jiang20a}, all of which have shown promising results. Yet, further reduction of sensitive attribute-related information in the learned embeddings is still an open problem in the area. Additionally, we observe that while attributes such as gender and age have been well-explored in this area, the notion of `attractiveness' which can cause considerable biases in human perception~\cite{talamas2016blinded}, has not been studied at all in the area of bias in FER.

In this paper, to address the notion of bias in FER, we propose a new solution that reduces sensitive attribute information from the learned embeddings of FER models. To this end, we utilize the kernel mean shrinkage (KMS) estimator \cite{JMLR:v17:14-195} to achieve a robust estimate of the kernel means of the distributions of the embeddings given the sensitive attributes (e.g., male vs. female). We then use this notion to measure the maximum mean discrepancy (MMD) distance between the distributions, and add this to our classifier loss to be minimized throughout the learning process. We call this the KMS loss.
% $\mathcal{L}_{KMS}$. 
Additionally, to further remove sensitive attribute information from the embeddings, we employ an adversarial mechanism.
% proposed in [?, ?]. 
The underlying concept of this adversarial mechanism is to decrease the accuracy of a classifier designed to identify sensitive attributes (e.g., male vs. female) from the embeddings, thereby minimizing the presence of such information throughout the learning process. It is worth mentioning that we show that the contribution of the proposed KMS loss 
% $\mathcal{L}_{KMS}$ 
is higher than the adversarial mechanism in promoting fairness. 
The proposed method ensures that when the FER model is trained, an accurate measure of the distance between the distributions of sensitive attributes is minimized, rendering that attribute unidentifiable for the model, hence improving fairness with respect to those attributes. We test the performance of our proposed method on two commonly used datasets in the area: RAF-DB \cite{li2017reliable} and CelebA \cite{liu2015faceattributes}. The results indicate that our devised method achieves strong results that outperform or are competitive to the state-of-the-art bias-dispelling solutions. Additionally, we define \textit{attractiveness} as a new sensitive attribute for which to reduce bias in FER models. We perform various experiments on this newly defined parameter and demonstrate that FER models are indeed biased toward more attractive faces. We also demonstrate that our solution for enhancing fairness in FER alleviates bias toward attractiveness as well.

In summary, our contributions in this paper are three-fold: 
\begin{enumerate}
    \item For promoting fairness in FER, we propose a new loss term based on KMS to reduce the MMD distance between the distributions of the representations learned by the FER model across different sensitive attribute groups. We demonstrate that our approach effectively estimates the misalignment between the distributions of the sensitive attribute groups.

    \item For the first time, we analyze the notion of \textit{attractiveness} as an important and sensitive attribute in FER models. Our experiments demonstrate that attractiveness is a sensitive attribute that can adversely impact the fairness of FER models. 

    \item We perform various experiments on two datasets and demonstrate that our method exhibits strong results on reducing bias toward gender, age, race, and attractiveness in FER, setting new state-of-the-art in several scenarios and achieving competitive performances in others. 
    
\end{enumerate}

The rest of the paper is organized as follows. In Section~\ref{section:related-work}, we present an overview of related work on bias in machine learning systems in general. The section then presents a literature review on measuring and reducing bias in facial analysis systems. In Section~\ref{section:proposed-method}, we present our proposed method. This is followed by Section~\ref{section:experiments}, which presents the experiment setup, and implementation details along with the results and ablation/sensitivity studies. Finally, Section~\ref{section:conclusion} concludes the paper with a summary and a discussion on possible future lines of inquiry.

\section{Related Work}\label{section:related-work}
In this section, we present an overview of the notion of bias in machine learning. We follow this up with a more focused review of the related work on bias and fairness in the context of facial analysis, which includes FER and facial recognition systems.

\subsection{Bias in Machine Learning}

Bias can occur due to various factors such as data imbalance, demographic distributions, and pre-processing steps \cite{10.1145/3457607}. Generally, bias-dispelling methods in machine learning can be applied in one of the following three stages~\cite{8682620}: pre-processing, in-processing, and post-processing. Pre-processing approaches usually apply some type of transformation on the data to reduce the amount of information regarding the sensitive attributes. In-processing methods aim to modify the underlying learning algorithms, for instance by modifying the loss terms. Post-processing methods are carried out after training. They usually rely on a test set which is not involved during the training phase to further process and modify the initially assigned labels.

In \cite{10.1145/3375627.3375865}, an innovative pre-processing technique is introduced which is aimed at reducing bias in machine learning applications. Specifically, the method augments the dataset by incorporating synthetically generated samples, thereby ensuring an equal distribution of instances across various groups within a given sensitive attribute. An example of an in-processing method is presented in \cite{DBLP:journals/corr/LouizosSLWZ15}, where the authors emphasize the utilization of MMD to promote fairness in deep generative models, such as variational autoencoders. In \cite{8682620}, a post-processing method is outlined. The method works by using a bias detector which prioritizes certain samples for processing by a fairness algorithm.

\subsection{Bias in Deep Facial Analysis}

Several methods have been proposed to study and address the issue of bias in deep facial analysis. In \cite{10.1007/978-3-030-65414-6_35}, the authors investigate gender, age, and race bias in the context of FER and show that FER systems can indeed be biased. To mitigate this, they employ adversarial and disentangled (DA) approaches to remove sensitive information from the generated embeddings and thus produce sensitive-attribute-agnostic embeddings. The effects of the data imbalance on the fairness of FER systems are studied in \cite{10.1145/3549865.3549904}. Specifically, the authors focus on gender bias and experimentally demonstrate that imbalanced datasets can alter the fairness of FER methods. The authors in \cite{9792455}, suggest the use of Continual Learning (CL) for bias mitigation in FER. Particularly, they use domain incremental CL for learning across different domains, each defined by sensitive attributes (e.g. female domain and male domain). They test different CL methods like elastic weight consolidation (EWC)~\cite{EWC}, synaptic intelligence (SI)~\cite{10.5555/3305890.3306093}, memory aware synapses (MAS)~\cite{10.1007/978-3-030-01219-9_9}, and naive rehearsal (NR)~\cite{hsu2018re}. 

Domain discriminative classification (DDC) is proposed in \cite{10.1145/2090236.2090255} to mitigate bias in FER methods. The authors argue that injecting the sensitive attribute information into the neural network would result in ``fairness through awareness''. 
% This information can be exploited by the neural network to promote fairness. 
In \cite{4bf20d93df4f4b4b8496afde33993ac0}, it is contended that DDC methods would make false classification boundaries within each class, and thus a domain independent classification (DIC) method is proposed. A new classifier is then trained for each sensitive attribute group. Strategic sampling (SS) \cite{9792455} methods highlight data imbalance as the cause of the bias, and thus address it by either re-sampling the original dataset to have equal samples across different sensitive attribute groups, or weighting the minority group. Although SS methods perform well on some sensitive attributes, recently in \cite{Wang_2019_ICCV}, it has been shown that balanced datasets are not enough to remove the bias.

The authors in \cite{Stoychev2022TheEO} study the effect of model compression on bias in the context of FER. They found inconclusive results for Extended Cohn-Kanade (CK+DB) and RAF-DB datasets. It appears that the effect of model compression on bias is still an open question in the area. Counterfactual data augmentation is used in \cite{10.1007/978-3-031-25072-9_16} for promoting fairness in FER systems. The authors perform data augmentation in three different stages, pre-processing, in-processing, and post-processing, and find that in-processing outperforms the others in terms of fairness. The authors in \cite{9710276} argue that one of the underestimated sources of bias in FER lies in dataset annotations. To remove the annotation biases, a network was trained with facial action units and triplet loss.

\section{Proposed Method}\label{section:proposed-method}

\subsection{Problem and Solution Overview}

% In the context of fairness and FER, 
Let us represent an FER training set as $\mathcal{D} = \{(x_i, y_i, z_i)\}_{i=1}^N$, where $x_i$, $y_i$, and $z_i$ denote the ith input face image, its corresponding expression label (happy, sad, etc.), and the associated sensitive attribute (e.g., old, young, child, etc.). For simplicity, we can assume that the training samples are independent and identically distributed (IID). Additionally, let $X$, $Y$, and $Z$ be the random variables associated with the input images, their labels, and the sensitive attributes. Let the distribution of $Z$ be defined over the set $\mathcal{Z}$. Also, let the cardinality of the set $\mathcal{Z}$ be equal to $\zeta$; for example, in the CelebA dataset, $\zeta=2$ for gender, which corresponds to male and female.

\begin{figure}[t] 
  \centering
  \includegraphics[width=0.5\linewidth]{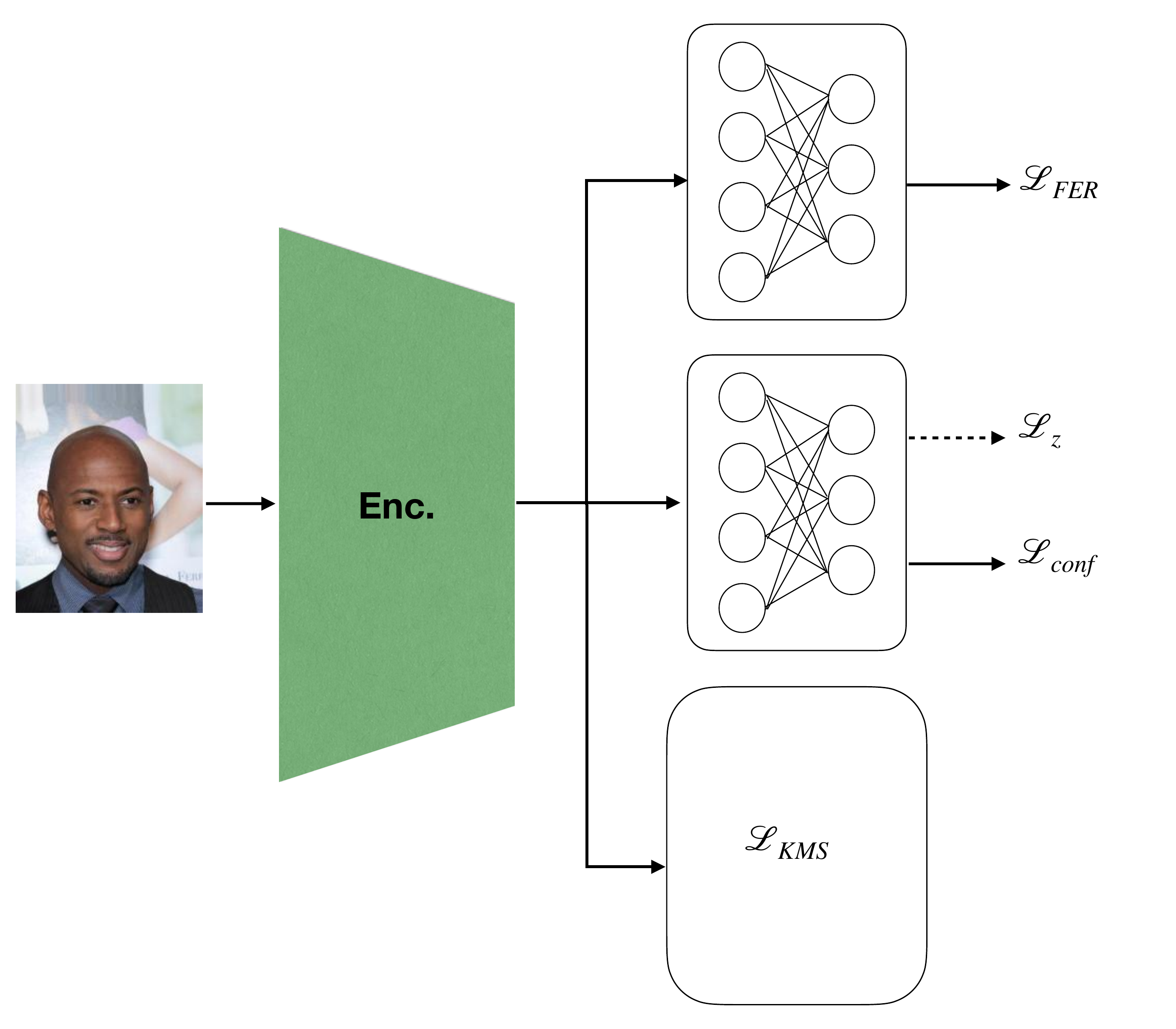}
  \caption{The proposed pipeline. $\mathcal{L}_{FER}$ update both the encoder and the corresponding classification head, $\mathcal{L}_{conf}$ and $L_{KMS}$ update the encoder, and $\mathcal{L}_{z}$, indicated by the dashed arrow, exclusively updates the classification head.}
\label{fig:pipeline}
\end{figure}

Accordingly, for a fair classifier, the optimum condition is:
% we should have:
\begin{equation}
  \mathbb{P}_{\hat{Y}}(\hat{y}|Z=z_i)=\mathbb{P}_{\hat{Y}}(\hat{y}|Z=z_j) \quad \forall z_i,z_j \in \mathcal{Z},
\end{equation}
% In the equation above, 
where $\hat{Y}$ represents the random variable associated with the output of the classifier, i.e., the predicted class. This criterion is also known as demographic parity. Bias-dispelling methods employ various strategies to achieve this parity criterion. 
A well-known approach is to use some form of adversarial learning to force the embeddings from which $\hat{y}$ is generated, to carry less information about $z_i$. 
However, it has been demonstrated that these embeddings may still contain such information, which can subsequently impact the fairness of the classifier. 
Therefore, in this paper, to further penalize the distance between the conditional distributions of the embeddings given the sensitive attribute $z_i$ , we calculate their relative MMD distance. MMD distance between two distributions is defined as the difference in the norm of their kernel means in the Hilbert space. In this paper, we calculate the kernel means of the distributions of the embeddings given the sensitive attributes using the KMS estimator. 
% Our main motivation to use the KMS estimator is that shrinkage estimators (like the KMS) show strong advantages over more naive estimators \cite{??}, especially this the case when the distributions are \textit{not perfectly normal}, for instance highly skewed or long-tailed \cite{??}. To further investigate and visualize this concept in the context of FER, we do the following: it can be easily proven that if the embeddings generated by an encoder come from an standard multivariate Gaussian distribution, then the distribution of their Mahalanobis distance \cite{??} follows a chi-square distribution. Thus, if one plots the quantiles of the Mahalanobis distance against the quantiles of the chi-square distribution, it should follow the $y=x$ line. This plot is named quantile-quantile (Q-Q) plot in the literature[]. The Q-Q plot of the embeddings of ResNet18 trained for FER is shown in Figure~\ref{fig:QQ}.  we observe that indeed the distribution is not normal . As a result, to better estimate the kernel mean, we propose to use the KMS estimator for calculating MMD. 
% we first compute the kernel mean of the distributions in the Hilbert space using the KMS estimator.
% Next, we calculate the MMD distance using the kernel means and 
The resulting MMD is then incorporated into the classifier's overall loss as $\mathcal{L}_{KMS}$. Additionally, we employ the adversarial strategy proposed in~\cite{Liu_2018_CVPR} to further ensure that the embeddings carry minimal information about the sensitive attribute. The adversarial strategy functions by steering the classifier, which aims to classify the sensitive attributes from the embeddings (e.g., detecting the age of the samples from the embeddings), towards a random classifier throughout the learning process. The proposed pipeline is illustrated in Figure~\ref{fig:pipeline}. As shown in this figure, the pipeline consists of a base feature extractor, an expression classification head, an adversarial component, and an MMD loss term, which enforces the network to extract fair embeddings. Following we describe the main components of our method in detail.

% \subsection{MMD-based Loss with KMS Estimator} 

% Embedding the probability measures in the Hilbert space has found many applications in machine learning [minimax estimation-Tolstikhin]. If $P$ denotes the probability measure which is defined over the $\mathcal{X}$ and $k$ is a positive-definite kernel, then P can be embedded into the Hilbert space, $H$, using $\mu_p=\int\Phi(x)P(x)dx= \E{X \sim P}\Phi(X)$ where $\Phi(x)=k(x, .)$ is a feature mapping from $\mathcal{X}$ to $\mathbb{R}$ and $\mu_p$ is called the kernel mean. Suppose that the space H is equipped with the norm $||.||_H$, then MMD defines the distance between the two probability measures P(x) and Q(y) as the norm of the difference between their corresponding kernel means:

\subsection{Kernel Mean Shrinkage Loss}

% Embedding probability distributions in the Hilbert space has found numerous applications in machine learning [minimax estimation-Tolstikhin]. 
Let $P$ denote a probability measure defined over $\mathcal{X}$, and $k$ be a positive-definite kernel. Then, $P$ can be embedded into the Hilbert space, $H$, using $\mu_p = \int\Phi(x)P(x)dx = \E{X \sim P}\Phi(X)$, where $\Phi(x) = k(x, .)$ is a feature mapping from $\mathcal{X}$ to $\mathbb{R}$, and $\mu_p$ is called the kernel mean. Assuming that the space $H$ is equipped with the norm $||.||_H$, the MMD defines the distance between two probability measures $P(x)$ and $Q(y)$ as the norm of the difference between their corresponding kernel means as
\begin{equation} \label{eq:MMD}
    MMD_{P, Q}=||\mu_P-\mu_Q||_{H}.
\end{equation}

\begin{figure}[t] 
  \centering
  \includegraphics[width=0.5\linewidth]{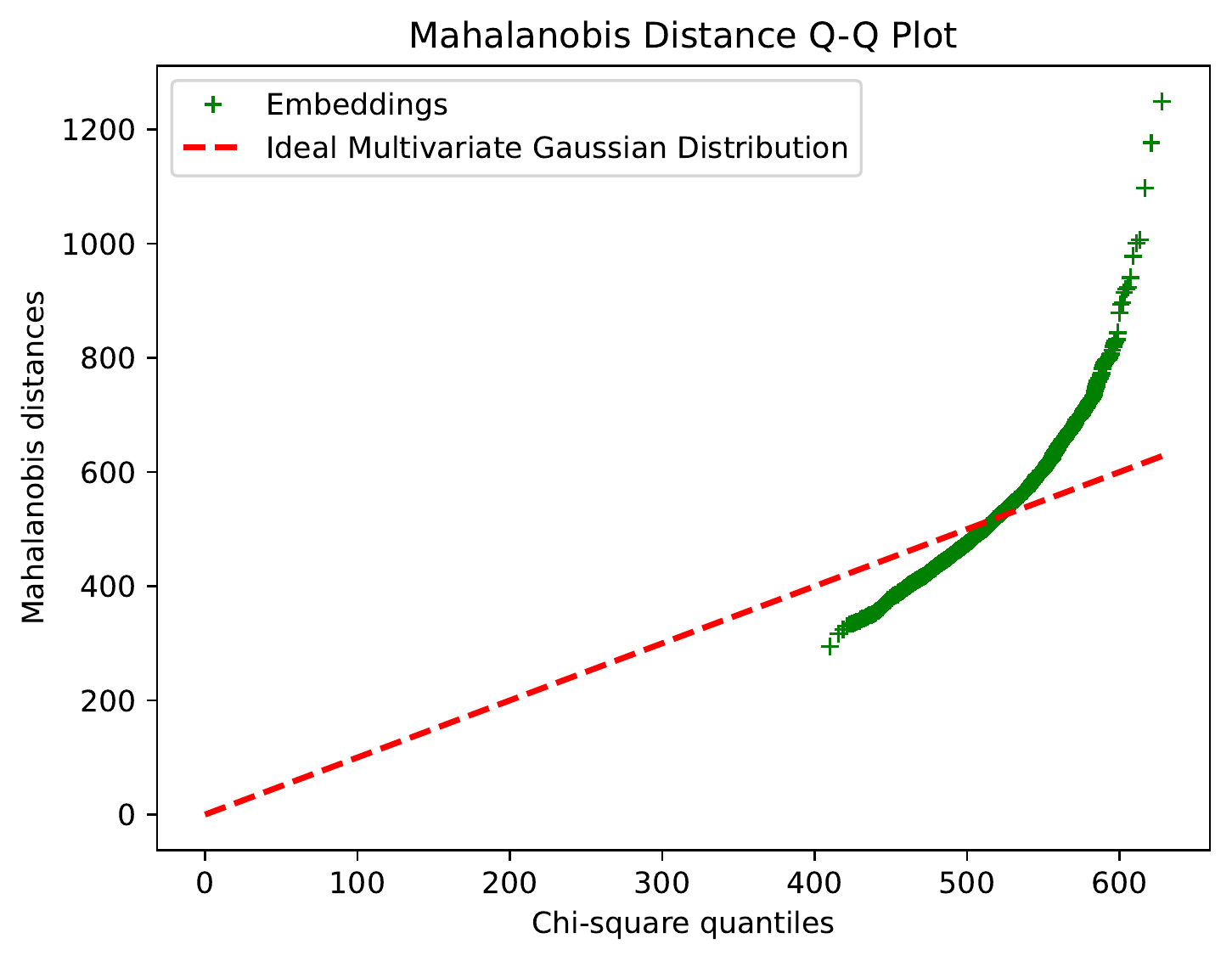}
  \caption{Quantile-quantile plot of the Mahalanobis distance for the output embeddings of the ResNet18, indicating that the embeddings do not follow a multivariate Gaussian distribution.}
\label{fig:QQ}
\end{figure}

In practice, estimates of $\mu_P$ and $\mu_Q$ are typically used. One of the most common estimates is the simple mean: $\hat{\mu}_P = \frac{1}{m}\Sigma_{i=1}^mk(x_i, .)$, where $m$ is the number of IID samples drawn from distribution $P$. Although widely adopted, it has been shown that better estimators exist, which can improve the simple mean estimator in terms of mean squared error \cite{JMLR:v17:14-195}. This is especially the case when the distributions are not perfectly normal, for instance highly skewed or long-tailed \cite{IKEDA201695}.  We can show that if a specific model (the encoder) generates data from a standard multivariate Gaussian distribution, the distribution of their Mahalanobis distance will follow a chi-square distribution. By plotting the quantiles of the Mahalanobis distance against the quantiles of the chi-square distribution, the resulting graph should follow $y=x$. This graph is called a quantile-quantile (Q-Q) plot. When we create a Q-Q plot using the data generated by the ResNet18 model trained for FER, we find that the distribution is indeed not normal (see Figure~\ref{fig:QQ}). As a result, to improve the estimation of the kernel mean, we suggest using the KMS estimator for calculating the MMD.

The KMS estimator, is defined by
% One such``better" estimator is KMS, defined as follows:
\begin{equation} \label{eq:KMS}
    \Tilde{\mu}_P = (1-\rho_P)\hat{\mu}_P,
\end{equation}
where, $\rho_P$ represents the \textit{shrinkage factor} and defined as
\begin{equation} \label{eq:rho-in-KMS}
    \rho_P=\frac{\hat{\Lambda}_P}{\hat{\Lambda}_P + ||\hat{\mu}_P||^2_H}.
\end{equation}
In this equation, $||\hat{\mu}_P||^2_H$ represents the squared norm of the $\hat{\mu}_P$ in $H$ which can be  calculated using the Riesz representation theorem \cite{adler2021hilbert}, also referred to as kernel trick. The kernel trick states that $<\Phi(x), \Phi(y)>_H=k(x, y)$. Thus, for the given $||\mu||^2_H$ we have:

\begin{equation} \label{eq:norm-mu}
\begin{split}
    ||\mu||^2_H=<\mu, \mu>_H & =<\E{X \sim P}\Phi(X),\E{X \sim P}\Phi(X)>_H\\
    & =\E{X \sim P}\E{X \sim P}<\Phi(X),\Phi(X)>_H\\
    & = \E{X \sim P}\E{X \sim P}k(x, x).
\end{split}
\end{equation}

The simple mean estimate of Equation~\ref{eq:norm-mu} gives 
$||\hat{\mu}_P||^2_H=\frac{1}{m^2}\Sigma_{i=1}^m\Sigma_{j=1}^mk(x_i,x_j)$. Additionally, in Equation~\ref{eq:rho-in-KMS}, $\hat{\Lambda}_P$ represents the risk associated with the KMS, and is defined as:
\begin{equation} \label{eq:risk-in-KMS}
    \hat{\Lambda}_P=\frac{\frac{1}{m}\Sigma_{i=1}^mk(x_i,x_i) - \frac{1}{m(m-1)}\Sigma_{i=1}^m\Sigma_{j\neq i}^mk(x_i,x_j) }{m}.
\end{equation}

By plugging Equation~\ref{eq:KMS} into Equation~\ref{eq:MMD} and using the kernel trick, we obtain the MMD through KMS, as:
\begin{equation} \label{eq:MMD-with-KMS}
\begin{split}
    MMD_{P,Q}^2 & = \frac{(1-\rho_P)^2}{m^2}\Sigma_{i=1}^m\Sigma_{j=1}^mk(x_i, x_j)\\ 
    & + \frac{(1-\rho_Q)^2}{n^2}\Sigma_{i=1}^n\Sigma_{j=1}^nk(y_i, y_j) \\
    & - \frac{2(1-\rho_P)(1-\rho_Q)}{mn}\Sigma_{i=1}^m\Sigma_{j=1}^nk(x_i, y_j),
\end{split}
\end{equation}
where $\rho_Q$ is defined similar to $\rho_P$ and $n$ indicates the number of IID samples drawn from the distribution $Q$. Using Equation~\ref{eq:MMD-with-KMS}, we define the KMS loss, $\mathcal{L}_{KMS}$, as follows:
\begin{equation} \label{eq:MMD-loss}
\mathcal{L}_{KMS} = \Sigma_{i=1}^{\zeta-1} \Sigma_{j=i+1}^{\zeta}MMD^{2}_{P_i, Q_j},
\end{equation}
which we use as one of the loss terms in our proposed method. In this equation, $P_i$ and $Q_j$ are the distributions of the output of the base feature extractor given $z_i$ and $z_j$, respectively. Recall, $\zeta$ is the cardinality of the set $\mathcal{Z}$, i.e., the number of elements in our sensitive attribute. %It is worth mentioning that since MMD is commutative, we define the $\mathcal{L}_{KMS}$ like Equation~\ref{eq:MMD-loss}.

\subsection{Adversarial Loss}
To further ensure that the base feature extractor generates embeddings that are agnostic to the sensitive attributes, we utilize an adversarial method inspired by \cite{Liu_2018_CVPR, 10.1007/978-3-030-65414-6_35}. As shown in Figure~\ref{fig:pipeline}, a classification head is applied to the output of the base feature extractor to classify the sensitive attributes. The idea is to aim to minimize the discriminability of these attributes.
% make the classification accuracy as low as possible. 
We use the following loss to train the base encoder:
\begin{equation} \label{eq:confusion-loss}
\mathcal{L}_{conf} = -\frac{1}{\zeta N}\Sigma_{j=1}^{N}\Sigma_{i=1}^{\zeta}log^{p(z_i|X=x_j)}.
\end{equation}
Here, $p(z_i|X=x_j)$ indicates the predicted probability that the output of the feature extractor for $x_j$ is assigned to the class $z_i \in \mathcal{Z}$. It should be noted that the minimum of $\mathcal{L}_{conf}$ occurs when the classification head predicts equal probabilities for all $z_i$ given $x_j$. 

However, it has been shown in prior works such as \cite{Liu_2018_CVPR} that the encoder in this scenario may learn a trivial solution that does indeed show poor discriminatory behavior, while the embeddings still contain strong attribute-related information.
% While this loss term does indeed promote fairness as we will later demonstrate in our ablation results, it has been shown in prior work such as \cite{??,???} that 
% However, the network can easily learn a trivial solution to not map φ(xi) to S even when φ(xi) contains sensitive information.
% classification head mitune iek trivial solution iad begire ke embedding hs ro be ie fazaie dg map kone ke too in faza deghate classification paiyn ahstesh amma still embedding ha havie etelaat dar morede sensitive attribute hastand. 
To avoid this trivial solution, an attribute predictive Cross Entropy loss is also added, as follows: % similar to \cite{??}.
\begin{equation} \label{eq:confusion-cross-entropy}
\mathcal{L}_{z} = -\frac{1}{\zeta N}\Sigma_{j=1}^{N}\Sigma_{i=1}^{\zeta}\mathbbm{1}[Z=z_i]log^{p(z_i|X=x_j)}.
\end{equation}
% One of the caveats of this approach is that 
% the head can learn a trivial??? solution and still
% the embeddings generated by the feature extractor could still contain information about the sensitive attribute. 
% Thus, the following cross-entropy loss is added to the method:
In the equation above, $\mathbbm{1}[Z=z_i]$ demonstrates the indicator function, $1$ if $Z=z_i$, and otherwise $0$. This equation acts as the adversary to Equation~\ref{eq:confusion-loss}. It should be noted that Equation~\ref{eq:confusion-loss} is used to train the encoder with the classification head fixed, while Equation~\ref{eq:confusion-cross-entropy} is only used to update the classification head.

\subsection{Total Loss}
To learn the FER-related representations which is the main goal of the model, we use $\mathcal{L}_{FER}$, which is the cross-entropy loss for the expressions, as follows:
\begin{equation} \label{eq:expression-cross-entropy}
\mathcal{L}_{FER} = -\frac{1}{\vartheta N}\Sigma_{j=1}^{N}\Sigma_{i=1}^{\vartheta}\mathbbm{1}[Y=y_i]log^{p(\hat{y}_i|X=x_j)}.
\end{equation}
Here, $\vartheta$ indicates the number of expressions in the dataset. Accordingly, we can now define the overall loss of our proposed method as follows:
\begin{equation} \label{eq:overall-loss}
\mathcal{L}_{total} = \gamma\mathcal{L}_{KMS} + \beta\mathcal{L}_{conf} + \mathcal{L}_z + \mathcal{L}_{FER}.
\end{equation}
In this equation, $\beta$ and $\gamma$ are two hyperparameters, which indicate the contribution of the adversarial and KMS losses, respectively. 

\section{Experiments}\label{section:experiments}

In this section, we first present the datasets used in our study. We then describe the pre-processing and data augmentation steps performed on the data. We then describe the implementation details of our method, followed by evaluation metrics. Next, for the first time, we propose \textit{attractiveness} as a new sensitive attribute and numerically and statistically show that there are indeed biases when considering this new attribute. Finally, we present the impact of our proposed method on the reduction of bias and compare our results to others in the area, along with ablation studies to evaluate the impact of different components and parameters of our method.

% the experiments and results pertaining to the analysis of bias induced by the notion of \textit{attractiveness} in FER models. We provide both statistical and experimental evidence demonstrating that FER models exhibit bias towards more attractive faces. Subsequently, to assess our proposed method for bias reduction, we conduct a series of experiments, comparing our approach's fairness and accuracy against several state-of-the-art models.

\subsection{Datasets}
\subsubsection{\textbf{CelebA}}
CelebA \cite{liu2015faceattributes} is a large-scale, real-world, and diverse facial dataset, encompassing 202,599 images from 10,177 unique subjects. The dataset features 40 distinct attribute annotations for each face image. In our study, we employ four attributes corresponding to facial expressions, age, gender, and attractiveness. Although CelebA does not cover the full spectrum of facial expressions, it does offer sensitive attribute information for face images, making it a widely-used benchmark in bias-mitigation research \cite{10.1007/978-3-030-65414-6_35}. To the best of our knowledge, no other dataset provides annotations for both facial expression recognition and attractiveness; thus, we rely solely on the CelebA dataset for the analysis of the bias caused by this attribute. Additionally, we utilize CelebA to evaluate our proposed method for bias reduction. The dataset offers official training, validation, and test sets.

\subsubsection{\textbf{RAF-DB}}
RAF-DB \cite{li2017reliable} is a large-scale, real-world facial dataset sourced from the internet. The dataset provides expression annotations (fear, sadness, disgust, anger, happiness, surprise, and neutral) and sensitive attribute annotations (race, age, and gender). We employ RAF-DB to assess our proposed method. To ensure a fair comparison other works in the area \cite{9792455}, we utilize a subset of RAF-DB containing 14,388 images featuring expression annotations, of which 11,512 images are utilized for training and the remaining for testing.

\subsection{Pre-processing and Augmentation}
For both datasets, images are aligned and cropped to ensure that faces appear approximately centered. Subsequently, we resize the images to $224 \times 224$ and normalize pixel values by dividing them by $255$. 
% Effective generalization in deep networks often requires large datasets; however, available facial expression datasets tend to have limited number of samples []. Consequently,
We also perform data augmentation to increase the size and diversity of our training sets. During training, each input image undergoes random rotation within the range of $-18\degree$ to $+18\degree$, horizontal flipping with a probability of $0.4$, and histogram equalization applied with a probability of $0.2$.

\subsection{Implementation Details}
To ensure a fair comparison with other works focusing on fairness \cite{9792455, 10.1007/978-3-030-65414-6_35}, we employ a ResNet18 \cite{7780459} as the base feature extractor. Hyperparameters $\beta$ and $\gamma$ in our total loss are tuned empirically 
% using random sampling from the uniform distribution within a specified range, 
with $\beta=0.14$ and $\gamma=0.17$. 

Our proposed model is implemented using TensorFlow \cite{abadi2016tensorflow} version 2.9, and training is conducted on two Nvidia RTX 2080 Ti GPUs. Furthermore, we utilize ADAM optimizer \cite{kingma2014adam} with a batch size of 64. The learning rate, first-momentum decay, and second-momentum decay in ADAM are set to 0.001, 0.9, and 0.99, respectively. The maximum number of epochs for each experiment is fixed at 50.

\subsection{Evaluation}
We report accuracy, precision, recall, F1 score, receiver operating characteristic curve (ROC), and the area under the ROC, also known as AUC. Additionally, we also report the fairness measure, which is widely used by literature in the area \cite{9792455, 10.1007/978-3-030-65414-6_35}, defined as:
% However, these metrics may not adequately represent the fairness of a classifier concerning sensitive attributes. Consequently, in addition to the aforementioned metrics, we employ a metric known as the fairness measure or simply fairness, as proposed in [] and []:
\begin{equation} \label{eq:fairness-measure}
\mathcal{F} = min(\frac{\Sigma_{c=1}^\vartheta p(\hat{y}|Y=c, Z=z, x)}{\Sigma_{c=1}^\vartheta p(\hat{y}|Y=c, Z=d, x)}) \quad \forall z \in \mathcal{Z}.
\end{equation}
Here, $d$ denotes the $z \in \mathcal{Z}$ with the highest mean classification accuracy (e.g., if female accuracy is higher than that of males, then $d=\text{female}$). In essence, $\mathcal{F}$ is the ratio of the lowest accuracy achieved for a given sensitive attribute (e.g., gender) to the highest accuracy attained for that specific attribute. Higher values of $\mathcal{F}$ indicate higher fairness in the classification task.

\begin{table}[t!]
\centering
\caption{Distribution of the CelebA dataset with regards to attractiveness as a sensitive attribute.}
\label{table:dist-celeba}
\begin{tabular}{c|c|c} 
\cline{2-3}
\multicolumn{1}{l|}{} & Smiling & Not Smiling  \\ 
\hline
Attractive            & 0.28    & 0.23         \\
Less Attractive       & 0.20    & 0.29         \\
\hline
\end{tabular}
\end{table}

\subsection{Analysis of the Bias towards Attractiveness}\label{subsection:attractiveness-analysis}
It has been well-observed that humans perceive and often discriminate favorably towards more attractive appearances \cite{talamas2016blinded}. This prompted us to investigate whether the same pattern holds true for neural networks, in this particular case, in the context of FER. 
% Due to the reasons discussed earlier, we employ CelebA dataset for analyzing the attractiveness bias. Our analysis begins with illustrating the distribution of the CelebA training set concerning the attractiveness sensitive attribute. 
As a first line of analysis, we explore the distribution of data with respect to attractiveness in the CelebA dataset, which contains a measurement for this attribute. Table~\ref{table:dist-celeba} displays this distribution. As evident from the table, the number of attractive samples exceeds that of less attractive ones. Next, we make two hypotheses in regards to attractiveness as a source of bias in FER systems: \\
\noindent (\textbf{1}) information pertaining to attractiveness are encoded in the learned embeddings;\\
% (similar to the fact that humans can perceive this attribute); 
\noindent (\textbf{2}) given that this attribute is both learned and the samples are imbalanced, classification results are biased towards more attractive samples.\\
% that, due to data imbalance, 
% leading the fully connected layer to utilize this information and make biased decisions. 
In this section, we test these hypotheses, and in the subsequent sections, we demonstrate that our method reduces such information and mitigates the bias in this regard.

\begin{figure}[t] 
  \centering
  \includegraphics[width=0.5\linewidth]{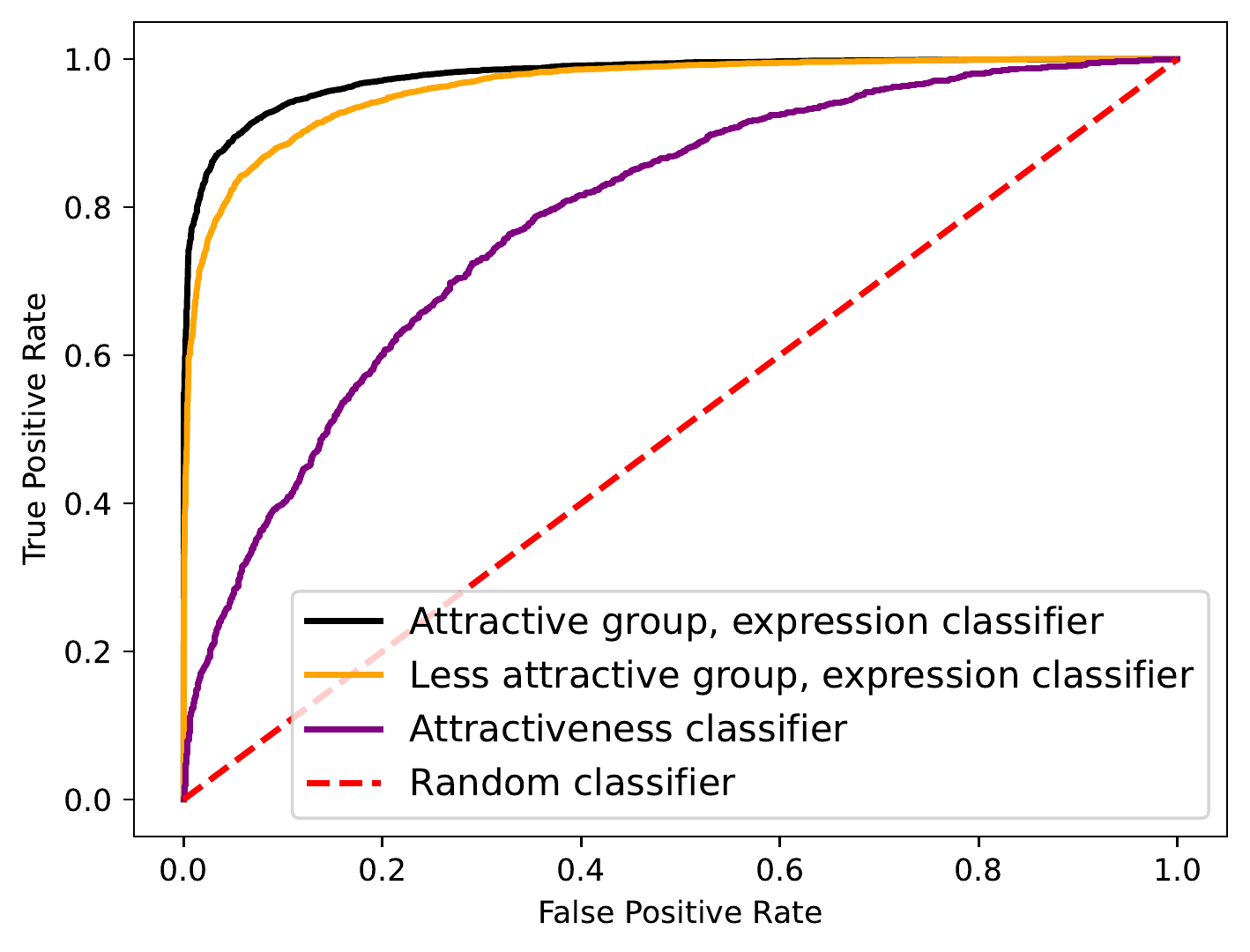}
  \caption{ROC curves for the expression classifier, as well as the attractiveness classifier.}
\label{fig:roc}
\end{figure}

To show that the embeddings generated by the feature extractor contain information about attractiveness, we perform the following steps, we first train the base feature extractor followed by a fully connected layer (classification head) for FER using the cross-entropy loss described in Equation~\ref{eq:expression-cross-entropy}, which henceforth we refer to as 
% We refer to this combination as 
`expression classifier'. Next, we remove the fully connected layer from the trained expression classifier and freeze the weights of the feature extractor and add a new, untrained fully connected layer. We then train the fully connected layer to predict the attractiveness of samples, using cross-entropy loss, while keeping the weights of the base feature extractor unchanged. We call this newly trained network the `attractiveness classifier'. Table~\ref{table:class-acc} presents the results. As shown in the table, the accuracy of the attractiveness classifier is $72\%$, which is significantly higher than chance level. Moreover, the AUC is 0.78, which also surpasses that of a random classifier by a large margin. We illustrate the ROC curve of the attractiveness classifier in Figure~\ref{fig:roc} alongside the ROC of the random classifier. Evidently, the ROC of the attractiveness classifier is above chance level, reinforcing the notion that the embeddings carry information about the attractiveness of the samples.

\begin{table*}[t!]
\centering
\caption{The results for the attractiveness and expression classifiers.}
\label{table:class-acc}
\begin{tabular}{c|c|c|c|c|c|c|c} 
\hline
\textbf{Classifier}                    & \textbf{Group}                       & \textbf{Accuracy} & \textbf{F1 Score} & \textbf{Precision} & \textbf{Recall} & \textbf{AUC} & \textbf{Fairness}                     \\ 
\hline
Attractiveness Classifier              & {\cellcolor[rgb]{0.753,0.753,0.753}} & 0.72              & 0.70              & 0.70               & 0.71            & 0.78         & {\cellcolor[rgb]{0.753,0.753,0.753}}  \\ 
\hline
\multirow{2}{*}{Expression Classifier} & Attractive                           & 0.92              & 0.93              & 0.92               & 0.93            & 0.97         & \multirow{2}{*}{0.95}                 \\
                                       & Less attractive                      & 0.88              & 0.87              & 0.84               & 0.90            & 0.96         &                                       \\
\hline
\end{tabular}
\end{table*}

\begin{figure}[t] 
  \centering
  \includegraphics[width=0.7\columnwidth]{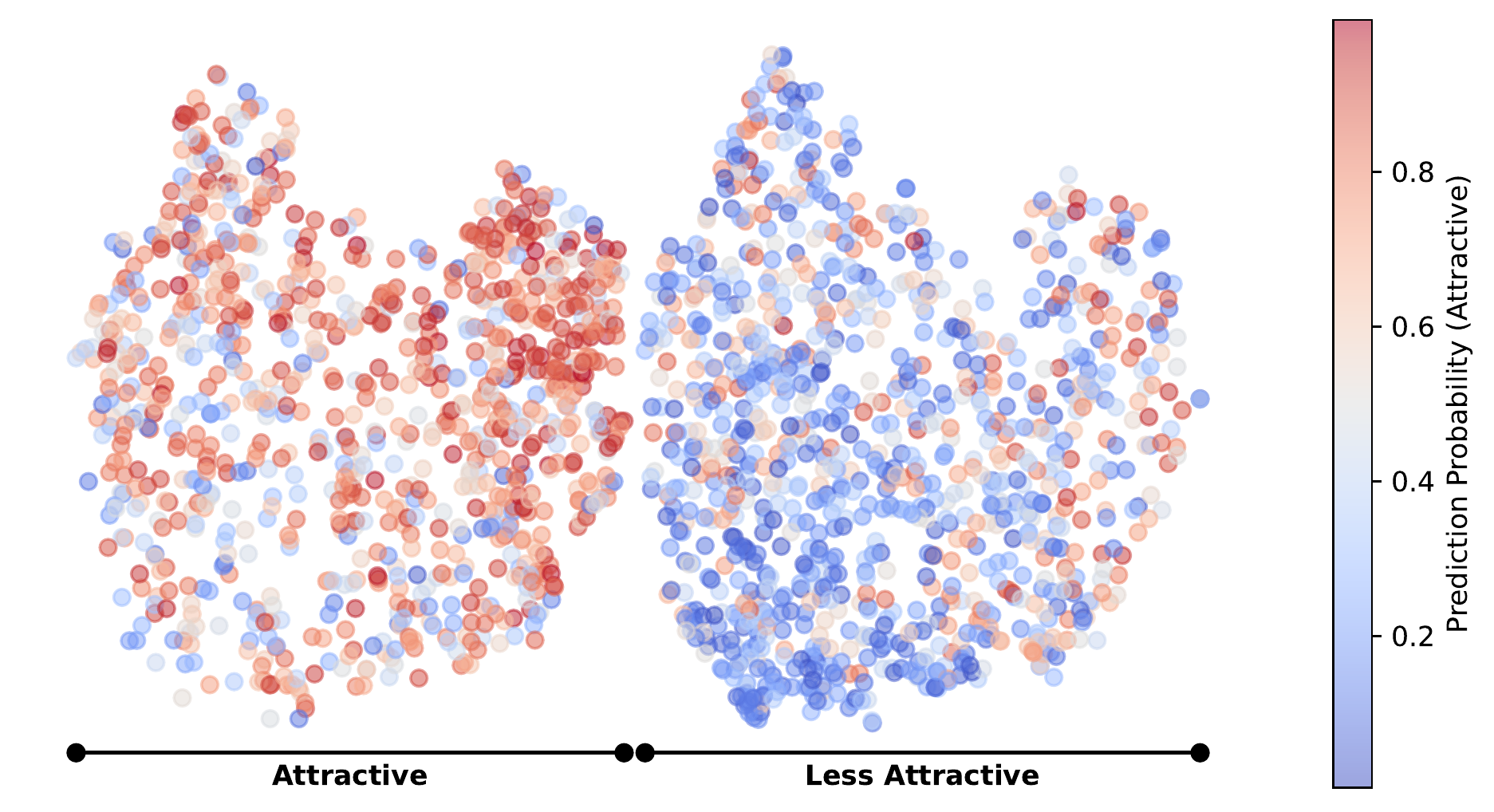}
  \caption{Visualization of the output of the attractiveness classifier. Here, t-SNE is used to depict the embeddings of different input samples, along with the probabilities of the model classifying them as attractive. The formation of a clear red cluster on the left indicates the ability of the model to identify features relating to attractiveness.}
\label{fig:tsne-attractiveness-classifier}
\end{figure}

To better visualize the output of the attractiveness classifier, we apply t-SNE \cite{JMLR:v9:vandermaaten08a} to map the embeddings generated by the feature extractor onto a 2-dimensional space. The results are illustrated in Figure~\ref{fig:tsne-attractiveness-classifier}. In this figure, the mapped embeddings with ground truth labels of `attractive' are displayed with on the left, while the `less attractive' samples are displayed on the right. 
% warmer (red) colors, while less attractive 
% on the left side, while the `less attractive' ones appear on the right side. 
Furthermore, the colors represent the probability of classifying a sample as attractive by the classifier 
% of correct classification generated by the fully connected layer, 
with warmer colors (red) indicating higher probabilities of a sample being labeled as attractive while colder colors (blue) denote the probabilities of samples being labeled as less attractive. As demonstrated, we observe that the model attains a strong ability to classify samples based on the attractiveness scores.
% in the samples with ground truths of `attractive', the model produces higher 
% warmer colors prevail, signifying higher correct classification probabilities, which in turn implies that the embeddings contain information about attractiveness. In the next section, we demonstrate that our method reduces such bias-distilling information.

\begin{table*}[t!]
\caption{T-test applied on the classification metrics for the attractive and less attractive groups as obtained from the expression classifier.}
\label{table:t-test}
\centering
\begin{tabular}{cccccc} 
\hline
\textbf{Measure} & \textbf{Accuracy}                       & \textbf{F1-score}                       & \textbf{Precision}                      & \textbf{Recall}                         & \textbf{AUC}                             \\ 
\hline
\textbf{P-value}          & 1.38 $\times$ 10\textsuperscript{-20} & 2.33 $\times$ 10\textsuperscript{-38} & 1.54 $\times$ 10\textsuperscript{-33} & 3.14 $\times$ 10\textsuperscript{-17} & 3.49 $\times$ 10\textsuperscript{-21}  \\
\hline
\end{tabular}
\end{table*}

\begin{table}[t!]
\caption{Accuracy and fairness scores for gender on the RAF-DB dataset.}
\label{table:gender-rafdb}
\centering
\begin{tabular}{c|c|c|c} 
\hline
\multirow{2}{*}{\textbf{Method}} & \multicolumn{2}{c|}{\textbf{Accuracy}} & \multirow{2}{*}{\textbf{Fairness}}  \\ 
\cline{2-3}
                        & \textbf{Male}          & \textbf{Female}        &                            \\ 
\hline
Offline~\cite{9792455}              & 0.72          & 0.75          & 0.95                       \\
Focal Loss~\cite{lin2017focal}           & 0.71          & 0.75          & 0.95                       \\
DDC~\cite{10.1145/2090236.2090255}                  & 0.71          & 0.74          & 0.96                       \\
DIC~\cite{4bf20d93df4f4b4b8496afde33993ac0}                  & 0.72          & 0.75          & 0.96                       \\
SS~\cite{9792455}                   & 0.72          & 0.76          & 0.95                       \\
DA~\cite{9792455}                  & 0.74          & 0.74          & \textbf{0.99}              \\
EWC~\cite{EWC}                  & 0.73          & 0.74          & 0.98                       \\
EWC-Online~\cite{schwarz2018progress}           & 0.73          & 0.75          & 0.97                       \\
SI~\cite{10.5555/3305890.3306093}                   & 0.73          & 0.73          & \textbf{0.99}              \\
MAS~\cite{10.1007/978-3-030-01219-9_9}                  & 0.74          & 0.75          & \textbf{0.99}              \\
NR~\cite{hsu2018re}                   & 0.73          & \textbf{0.79} & 0.92                       \\
Ours                 & \textbf{0.75} & 0.75          & \textbf{0.99}              \\
\hline
\end{tabular}
\end{table}

Considering the second hypothesis, we present the results of the expression classifier for both attractive and less attractive groups in Table~\ref{table:class-acc}. As demonstrated in the table, there are differences of 4\% and 6\% between the accuracy and F1 scores of the two groups, respectively. Furthermore, the fairness score for the expression classifier is 0.95. These findings illustrate that the expression classifier is biased towards the attractive group, exhibiting better expression recognition for this group. To provide a clearer visualization of the differences between the attractive and less attractive groups, we plot the ROC curves for both groups in Figure~\ref{fig:roc}. As depicted in the figure, the yellow curve, corresponding to the less attractive group lies below the black curve, which corresponds to the attractive group, confirming the existence of bias towards the attractive group.

Lastly, to statistically evaluate whether the difference between the results of the attractive and less attractive groups as per the expression classifier is significant or not, we employ a two-sample t-test. For this purpose, we randomly select 500 samples from both attractive and less attractive groups and calculate the classification metrics for each group. We repeat this process 100 times and apply the two-sample t-test on the computed metrics. The null hypothesis assumes that there is no significant difference between the average of the metrics, while the alternative hypothesis suggests otherwise. Table~\ref{table:t-test} presents the obtained p-values. Assuming an extremely strong significance level of $0.001$, 
% $$10^{-5}$, 
we reject the null hypothesis for all metrics, indicating significant differences between the metrics calculated for the attractive and less attractive groups. This experiment further verifies the possibility of bias towards the attractive group in FER systems.

\begin{table}[t!]
\centering
\caption{Accuracy and fairness scores for gender on the CelebA dataset.}
\label{table:gender-celeba}
\begin{tabular}{c|c|c|c} 
\hline
\multirow{2}{*}{\textbf{Method}} & \multicolumn{2}{c|}{\textbf{Accuracy}} & \multirow{2}{*}{\textbf{Fairness}}  \\ 
\cline{2-3}
                                 & \textbf{Male} & \textbf{Female}        &                                     \\ 
\hline
Baseline~\cite{10.1007/978-3-030-65414-6_35}                      & 0.91          & 0.93                   & 0.97                                \\
Attribute-aware~\cite{10.1007/978-3-030-65414-6_35}               & 0.91          & 0.93                   & 0.97                                \\
Disentangled~\cite{10.1007/978-3-030-65414-6_35}                     & 0.91          & 0.93                   & 0.97                                \\
Ours                             & \textbf{0.93}          & \textbf{0.94}                   & \textbf{0.99}                                \\
\hline
\end{tabular}
\end{table}

\subsection{Performance of Our Method}

We train the FER system comprising the ResNet18 feature extractor and expression classification head illustrated in Figure~\ref{fig:pipeline} using the loss proposed in Equation~\ref{eq:overall-loss}. We employ both RAF-DB and CelebA datasets as discussed earlier. Considering gender as the sensitive attribute, we report the obtained fairness measure and accuracy in Table~\ref{table:gender-rafdb} for the RAF-DB dataset. We observe that our method achieves an accuracy of 75\% for both male and female groups, surpassing other state-of-the-art methods in the male group, while NR~\cite{hsu2018re} achieves a higher accuracy for the female group. However, it should be noted that we achieve a fairness score of 99\% which is considerably higher than~\cite{hsu2018re} and equal to the state-of-the-art. Moreover, our method is one of the few that achieves equal performance across both groups.
% which is slightly higher than ours. In terms of the fairness measure, we achieve the value of 0.99, which is also the current state-of-the-art and matches the performance of DA [], SI [], and MAS []. 
For the CelebA dataset, we report the results in Table~\ref{table:gender-celeba}. In this experiment, we observe that our proposed method outperforms the current state-of-the-art in terms of the accuracy for both male and female groups, as well as the fairness measure, setting a new state-of-the-art. Moreover, the performance of our approach across both groups is closer in comparison to the prior works.

Regarding the sensitive attribute of race, we present the results in Table~\ref{table:race-rafdb} for the RAF-DB dataset. In terms of accuracy for Caucasian, African-American, and Asian groups, our method achieves 78\%, making it and SI~\cite{10.5555/3305890.3306093} the only methods that obtain similar accuracies across all race groups. 
% This makes them as the fairest classifiers. 
Furthermore, the fairness measure for our proposed method is 0.99, which is equal to the current state-of-the-art for race as the sensitive attribute. Recall, CelebA does not provide race-related ground truth labels.

\begin{table}[t!]
\caption{Accuracy and fairness scores for race on the RAF-DB dataset.}
\label{table:race-rafdb}
\centering
\setlength
\tabcolsep{4pt}
\begin{tabular}{c|c|c|c|c} 
\hline
\multirow{2}{*}{\textbf{Method }} & \multicolumn{3}{c|}{\textbf{Accuracy }}                         & \multirow{2}{*}{\textbf{Fairness }}  \\ 
\cline{2-4}
                                  & \textbf{Caucasian} & \textbf{Af.-Am.} & \textbf{Asian} &                                      \\ 
\hline
Offline~\cite{9792455}                        & 0.74               & 0.76                      & 0.76           & 0.93                                 \\
Focal Loss~\cite{lin2017focal}                     & 0.73               & 0.75                      & 0.75           & 0.97                                 \\
DDC~\cite{10.1145/2090236.2090255}                            & 0.72               & 0.73                      & 0.74           & 0.97                                 \\
DIC~\cite{4bf20d93df4f4b4b8496afde33993ac0}                            & 0.74               & 0.76                      & 0.77           & 0.96                                 \\
SS~\cite{9792455}                             & 0.74               & 0.75                      & 0.76           & 0.97                                 \\
DA~\cite{9792455}                             & 0.75               & 0.76                      & 0.70           & 0.91                                 \\
EWC~\cite{EWC}                            & \textbf{0.79}      & 0.78                      & 0.79           & \textbf{0.99}                        \\
EWC-Online~\cite{schwarz2018progress}                     & 0.77               & 0.78                      & 0.78           & \textbf{0.99}                        \\
SI~\cite{10.5555/3305890.3306093}                             & 0.78               & 0.78                      & 0.78           & \textbf{0.99}                        \\
MAS~\cite{10.1007/978-3-030-01219-9_9}                            & 0.78               & 0.77                      & 0.78           & 0.97                                 \\
NR~\cite{hsu2018re}                             & 0.78               & \textbf{0.79}             & \textbf{0.80}  & \textbf{0.99}                        \\
Ours                           & 0.78               & 0.78                      & 0.78           & \textbf{0.99}                        \\
\hline
\end{tabular}
\end{table}

The results of the proposed method for age as the sensitive attribute on RAF-DB and CelebA datasets are reported in Tables~\ref{table:age-rafdb} and~\ref{table:age-celeba}, respectively. The proposed method achieves the highest accuracy for the 4-19, 20-39, 40-69, and 70+ age groups. However, in the 0-3 age group, the baseline~\cite{10.1007/978-3-030-65414-6_35} method achieves the highest accuracy. Furthermore, our proposed method attains a fairness score of $0.84$, which sets a new state-of-the-art. Regarding the CelebA dataset, our method acquires a fairness score of 0.98, which equals the current state-of-the-art.

\begin{table}[t!]
\caption{Accuracy and fairness scores for age on the RAF-DB dataset.}
\setlength
\tabcolsep{2pt}
\label{table:age-rafdb}
\centering
\begin{tabular}{c|c|c|c|c|c|c} 
\hline
\multirow{2}{*}{\textbf{Method }} & \multicolumn{5}{c|}{\textbf{Accuracy }}                                         & \multirow{2}{*}{\textbf{Fairness }}  \\ 
\cline{2-6}
                                  & \textbf{0-3}  & \textbf{4-19} & \textbf{20-39} & \textbf{40-69} & \textbf{70+}  &                                      \\ 
\hline
Baseline~\cite{10.1007/978-3-030-65414-6_35}                       & \textbf{0.80} & 0.61          & 0.74           & 0.73           & 0.60          & 0.75                                 \\
Attribute-aware~\cite{10.1007/978-3-030-65414-6_35}                & 0.71          & 0.63          & 0.75           & 0.74           & 0.54          & 0.71                                 \\
Disentangled~\cite{10.1007/978-3-030-65414-6_35}                   & 0.65          & 0.69          & 0.76           & 0.72           & 0.62          & 0.81                                 \\
Ours                              & 0.69          & \textbf{0.72} & \textbf{0.78}  & \textbf{0.75}  & \textbf{0.66} & \textbf{0.84}                        \\
\hline
\end{tabular}
\end{table}

\begin{table}[t!]
\centering
\caption{Accuracy and fairness scores for age on the CelebA dataset.}
\label{table:age-celeba}
\begin{tabular}{c|c|c|c} 
\hline
\multirow{2}{*}{\textbf{Method}} & \multicolumn{2}{c|}{\textbf{Accuracy}} & \multirow{2}{*}{\textbf{Fairness}}  \\ 
\cline{2-3}
                                 & \textbf{Old}  & \textbf{Young}         &                                     \\ 
\hline
Baseline~\cite{10.1007/978-3-030-65414-6_35}                      & 0.91          & 0.93                   & 0.97                                \\
Attribute-aware~\cite{10.1007/978-3-030-65414-6_35}               & 0.91          & 0.93                   & 0.97                                \\
Disentangled~\cite{10.1007/978-3-030-65414-6_35}                  & 0.91          & 0.93                   & \textbf{0.98}                       \\
Ours                             & \textbf{0.92} & \textbf{0.94}          & \textbf{0.98}                       \\
\hline
\end{tabular}
\end{table}

We also test our proposed method on attractiveness as the sensitive attribute. Since, this is the first work to study this attribute in the context of bias, there are no prior papers that report fairness measures in this context. We therefore compare our method against the expression classifier discussed earlier (see Section~\ref{subsection:attractiveness-analysis}) as a baseline. The results are reported in Table~\ref{table:attractiveness-celeba} for the CelebA dataset. As it is shown in the table, our proposed method not only increases FER performance for both attractive and less attractive groups, it also promotes fairness at the same time. Furthermore, we evaluate whether our proposed method reduces the amount of attractiveness-related information in the learned embeddings generated by the base feature extractor. To this end, we use a setup similar to the attractiveness classifier; however, this time, we train the model using Equation~\ref{eq:overall-loss}. We then remove the FER classification head, freeze the weights of the base feature extractor, and train a new head (FC layer) to classify attractiveness. We observe that, in comparison to the attractiveness classifier which achieved an accuracy of 72\%, the accuracy of the model trained using our proposed approach drops to 54\%. This clearly demonstrates that the embeddings carry less information about attractiveness.

\begin{table}[t!]
\centering
\setlength
\tabcolsep{2pt}
\caption{Accuracy and fairness for attractiveness on CelebA dataset}
\label{table:attractiveness-celeba}
\begin{tabular}{c|c|c|c} 
\hline
\multirow{2}{*}{\textbf{Method}} & \multicolumn{2}{c|}{\textbf{Accuracy}}         & \multirow{2}{*}{\textbf{Fairness}}  \\ 
\cline{2-3}
                                 & \textbf{Attractive} & \textbf{Less Attractive} &                                     \\ 
\hline
Expression classifier                        & 0.92                & 0.88                     & \multicolumn{1}{c}{0.95}            \\
Ours               & \textbf{0.95}       & \textbf{0.93}            & \multicolumn{1}{c}{\textbf{0.98}}   \\
\hline
\end{tabular}
\end{table}

\begin{table}[t!]
\centering
\caption{Ablation study. The drops in the fairness are provided when the specific terms from the loss function is removed.}
\label{table:ablation}
\begin{tabular}{c|c|c} 
\hline
\multirow{2}{*}{\textbf{Removed Term}} & \multicolumn{2}{c}{\textbf{Drop in the Fairness}}  \\ 
\cline{2-3}
                                      & \textbf{Gender} & \textbf{Attractiveness}          \\ 
\hline
$\mathcal{L}_{KMS}$  & 0.02            & 0.02                      \\
$\mathcal{L}_{z}+\mathcal{L}_{conf}$       & 0.01            & 0.01                             \\
$\mathcal{L}_{KMS}+\mathcal{L}_{z}+\mathcal{L}_{conf}$  & 0.04            & 0.03                             \\
\hline
\end{tabular}
\end{table}

\subsection{Ablation Study}
We perform ablation studies to evaluate the contribution of each term in the proposed total loss (see Equation~\ref{eq:overall-loss}). This is accomplished by removing one specific term from the loss function and measuring the drop in fairness. We use CelebA dataset and report the results in Table~\ref{table:ablation}. As it is shown in the table, the ablation of each term results in a drop in performance. Moreover, we observe that the proposed $\mathcal{L}_{KMS}$ term is the most effective term in the loss function in comparison to the other terms as its removal results in the highest drop in fairness.

\subsection{Limitations}
One of the potential limitations of our work is its computational complexity. Since the proposed $\mathcal{L}_{KMS}$ involves computing the output of the more advanced KMS estimator instead of the sample mean, it requires more training time and computational resources. Given that ground truth labels for attractiveness have only been provided in the CelebA dataset, this was our only means of studying this sensitive attribute in this paper.
% Additionally, due to the availability of both expression and attractiveness labels, we only have used CelebA dataset to analyze the attractiveness bias. 
For more concrete analysis, more datasets with different setups should be eventually collected and considered. Lastly, the notion of attractiveness is subjective and can vary across cultures and individuals. As a result, our approach to reducing bias in this context is limited to the dataset that was available to us at the time of performing this study.%  and handling this attribute may not fully capture the complexity and diversity of this concept.

\section{Conclusion and Future Work}\label{section:conclusion}
This paper presented a novel method to address the notion of bias in FER systems. To this end, our devised method minimizes the sensitive attribute information like gender, existing in the embeddings generated by the feature extractor (encoder). Motivated by the fact that these embeddings do not follow multivariate Gaussian distribution, we use the more advanced KMS estimator for calculating the embeddings' kernel means in the Hilbert space and subsequently we calculate the MMD distance using the estimated kernel means. The MMD distance calculated using the KMS estimator is incorporated in the total loss of the network. Additionally, we use adversarial term in the proposed loss to further ensure that embeddings do not carry information about sensitive attributes. We, also, highlight \textit{attractiveness} as an important sensitive attribute and for the first time analyze the resulting bias in the context of FER. Both experimentally and statistically, we show that the FER systems are biased toward more attractive faces. Our bias-distilling method not only reduces the attractiveness bias but also it sets new state-of-the-art fairness scores in the case of gender and age sensitive attributes. 

Future work could include the theoretical analysis of the KMS estimator to demonstrate the reason behind its superiority in promoting fairness, and whether potential limitations exist in its usage. Lastly, the notion of robustness in the context of our work can be examined, for instance by considering noisy and partially occluded samples. Such experiments could determine how our method deals with such challenging scenarios in comparison to others.

\section{Acknowledgements}
This work was funded by Irdeto Canada Corporation and the Natural Sciences and Engineering Research Council of Canada (NSERC).

%Bibliography
\bibliographystyle{unsrt}  
\bibliography{references}

\end{document}